\documentclass{article}

\usepackage[preprint]{neurips_2023}

\usepackage[utf8]{inputenc}
\usepackage[T1]{fontenc}
\usepackage{hyperref}
\usepackage{url}
\usepackage{booktabs}
\usepackage{amsfonts}
\usepackage{amsmath,amssymb,mathtools}
\usepackage{bm}
\usepackage{nicefrac}
\usepackage{microtype}
\usepackage{graphicx}
\usepackage{multirow}
\usepackage{subcaption}
\usepackage{xcolor}
\usepackage{enumitem}

\graphicspath{{figures/}}

\DeclareMathOperator{\sigmoid}{sigmoid}
\newcommand{\R}{\mathbb{R}}
\newcommand{\softmax}{\mathrm{softmax}}

\title{Spatially Grounded Concept Bottleneck Models via Part-Factorized Attention}

\author{%
  Dhanesh Ramachandram\\
  \texttt{dhanesh.ramachandram@vectorinstitute.ai}\\
}

\begin{document}

\maketitle

\begin{abstract}
Concept bottleneck models (CBMs) predict a layer of human-named attributes before predicting a class, which makes their decisions auditable. On fine-grained recognition tasks, though, the concept heads are usually free to attend anywhere in the image, so a head named for one body region can be satisfied by evidence on another, and the model reaches the right answer for the wrong reason. We propose a part-factorized CBM (PF-CBM) that removes this freedom by construction. A frozen DINOv3 vision transformer feeds a set of part queries, each tied by name to a specific anatomical region through a fixed concept-to-part map, while whole-object attributes such as size and shape are handled separately by a query with no spatial prior, since they are not anchored to any single body part. A learnable Gaussian prior over patch locations, initialized from average keypoint positions, keeps the part queries from collapsing onto the same evidence. On its own this prior spreads the queries apart but does not reliably land them on the correct anatomy. What closes that gap is a lightweight alignment loss that nudges each part query toward its keypoint, and the central finding of this paper is how little of that supervision is required. Aligning on well under one percent of the training images already moves pointing accuracy from near-chance to roughly three-quarters of what full keypoint supervision achieves, and the gains continue, more slowly, as more annotated images are added. Classification accuracy on CUB-200-2011 barely moves across this entire range and stays within a point of a fully supervised baseline whether the model sees no keypoints at all or every one of them. Grounding a CBM's attention to the right evidence turns out to be nearly free in accuracy and cheap in annotation, provided the model has the right inductive bias to make efficient use of that small amount of supervision.
\end{abstract}

\section{Introduction}
\label{sec:intro}

Fine-grained recognition separates classes that look alike, such as 200 bird species, by reasoning about small localized differences: the curve of a bill, the pattern of a wing bar, the color of a throat. Datasets such as CUB-200-2011 \citep{wah2011cub} encode this knowledge as binary part-attribute labels (eg, \emph{has\_bill\_shape::all-purpose}, \emph{has\_throat\_color::white}). That structure suits a concept bottleneck model (CBM) \citep{koh2020concept}, which predicts attributes first and the class from those attributes second, exposing the middle layer as a human-readable explanation. Interpretability of this kind is imperative in high-stakes settings, where a practitioner needs to know not only the prediction but the evidence behind it, and to correct that evidence when it is wrong.

Although the promise is attractive, it is hard to keep in practice. A plain CBM over a strong backbone learns the 312 attribute heads as 312 independent spatial attenders, each free to point anywhere. Because attributes and classes are strongly correlated in a fine-grained dataset, the training loss can be satisfied by a throat-color head that attends to a wing patch, provided that patch happens to predict the throat attribute well enough. The bottleneck stays numerically interpretable (ie, one sigmoid per attribute), but it is spatially ungrounded. The concept layer no longer indicates which region the evidence came from, and test-time interventions act through whichever patch a head latched onto rather than through the named anatomy. This is the spatial grounding gap that motivates our work.

We studied a part-factorized concept bottleneck model, PF-CBM, that makes spatial grounding a property of the forward graph. The model is built on a frozen DINOv3 vision transformer \citep{oquab2024dinov3}, and we contribute four mechanisms:
\begin{enumerate}[leftmargin=1.5em,itemsep=2pt,topsep=2pt]
  \item a learned foreground gate over DINOv3 patch features that suppresses background regions inside the part attention, injected additively in log space so that it is differentiable everywhere;
  \item a slot-to-part routing scheme that aligns a small set of part queries with named anatomical parts and routes each of the $293$ part-specific attributes, through a fixed concept-to-part map, to read only from the part token its name implies;
  \item a dedicated global token attention pathway that handles the $19$ whole-object attributes (ie, \emph{has\_size} and \emph{has\_shape}): a single learnable global query cross-attends over all foreground patches with no Gaussian spatial prior, because these attributes describe whole-object properties that are not tied to any anatomical region;
  \item a learnable Gaussian spatial prior over expected part locations that breaks the permutation symmetry among part queries and seeds their means, paired with a centroid alignment loss that supplies anatomical grounding from keypoint annotation on a small fraction of images (as few as $0.5\%$, about $27$).
\end{enumerate}

The rest of the paper is organized as follows. Section~\ref{sec:related} reviews CBMs, foundation-model-assisted concept generation, spatially grounded concept learning, and slot attention. Section~\ref{sec:method} defines the forward graph and the training objective. Section~\ref{sec:experiments} describes the dataset, baselines, metrics, and implementation. Section~\ref{sec:results} reports the main results, ablations, the sequential-versus-joint comparison, and qualitative grounding. Section~\ref{sec:conclusion} states limitations and future directions.

\section{Background and Related Work}
\label{sec:related}

\paragraph{Concept bottleneck models.}
Koh et al.\ \citep{koh2020concept} reintroduced CBMs as a two-stage model that maps an image to a vector of human-named concepts and then maps that vector linearly to the class, supporting test-time concept intervention. Two follow-up strands shaped the field. The first weakens the labeling requirement. Post-hoc CBMs \citep{yuksekgonul2023pcbm} retrofit a concept layer onto a pretrained backbone using textual concept embeddings, and label-free CBMs \citep{oikarinen2023lfcbm} source both concepts and supervision from CLIP. The second strand questions whether the concept layer captures the intended semantics. Mahinpei et al.\ \citep{mahinpei2021leakage} documented concept leakage, in which soft concept scores encode unintended task information, and concept embedding models \citep{zarlenga2022cem} trade scalar concepts for higher-dimensional embeddings to recover accuracy. Our PF-CBM keeps scalar, named concepts and instead constrains where each concept reads from.

\paragraph{Foundation-model-assisted concept generation.}
Several methods automate concept-set construction. LaBo \citep{yang2023labo} uses a language model to propose a large candidate concept space and a submodular selection to pick discriminative, diverse concepts that are aligned to images through CLIP. Label-free CBMs \citep{oikarinen2023lfcbm} similarly remove manual concept annotation. These methods address concept supply; in contrast, our work takes the CUB attribute set as given and addresses the spatial grounding of those concepts.

\paragraph{Spatially grounded concept learning.}
A recent line asks where in the image a concept comes from. VLG-CBM \citep{vlgcbm2024neurips} pairs each concept with a bounding box from an open-vocabulary detector to prevent the model from reporting concepts that are not present. DCBM \citep{dcbm2025icml} replaces boxes with regions from a segmentation foundation model. Most relevant is DOT-CBM \citep{xie2025dotcbm}, which models concept prediction as an optimal-transport problem between image patches and concepts. It computes a patch-to-concept assignment through Sinkhorn iterations and uses a saliency map together with concept-label statistics as transportation priors, with orthogonal-projection regularizers that disentangle patch and concept features. DOT-CBM uses a frozen DINOv2 ViT-L/14 image encoder and a CLIP text encoder, and reports $85.39\%$ top-1 on CUB-200-2011. Our PF-CBM shares the goal of fine-grained visual-concept localization and the use of a frozen DINO backbone, but it differs in two ways. DOT-CBM aligns concepts to patches softly, through a learned transport plan, so any concept may in principle draw on any patch; our PF-CBM instead routes each concept to a single anatomical part token through a fixed map, so grounding is a structural guarantee rather than the outcome of an optimization. DOT-CBM uses a saliency prior to discourage background shortcuts, which is comparable in spirit to our foreground gate, while our Gaussian spatial prior additionally fixes which part query covers which region. The two approaches therefore sit on opposite ends of a soft-versus-structural spectrum for grounding the same kind of concept layer.

\paragraph{Slot attention and part-based models.}
Slot attention \citep{locatello2020slot} introduced a competitive cross-attention over learnable slots for unsupervised scene decomposition, and locality-biased variants such as Spotlight Attention \citep{kakogeorgiou2023spotlight} add a spatial prior that pulls each slot toward a compact region. Part-discovery methods build on frozen DINO features in the same spirit. PDiscoNet \citep{klis2023pdisconet} learns part heatmaps from class labels under compactness and equivariance constraints, and PDiscoFormer \citep{aniraj2024pdiscoformer} relaxes the compactness prior with a total-variation term and reports state-of-the-art unsupervised part discovery on CUB. PDiscoFormer is the closest architectural neighbor, a frozen DINO backbone with part slots that cross-attend over patches. It outputs part maps and a classifier over their pooled features, with no concept layer and no concept intervention. Prototype methods such as ProtoPNet \citep{chen2019protopnet} and ProtoViT \citep{ma2024protovit} give part-level explanations without an intermediate named-concept layer. The additive log-space bias used by our spatial prior follows ALiBi \citep{press2022alibi}, which showed that a logit-level bias can shape attention as effectively as positional embeddings. Our PF-CBM draws on each of these directions, connecting frozen DINO features, part-level cross-attention, and a named concept bottleneck through a fixed concept-to-part routing that ties spatial grounding directly to concept prediction.

\section{Method}
\label{sec:method}

\subsection{Overview and notation}
\begin{figure}[t]
  \centering
  \includegraphics[width=0.6\linewidth]{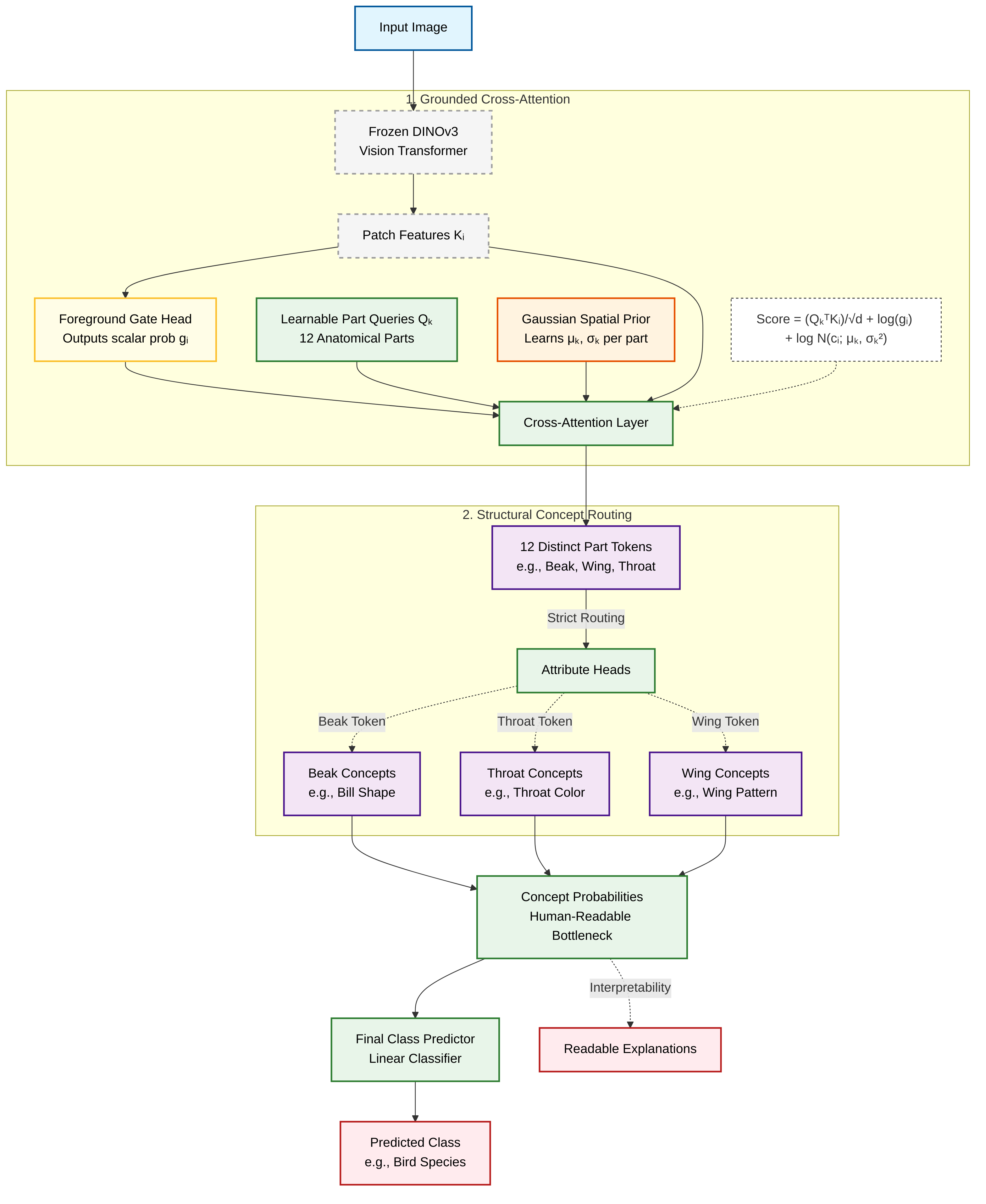}
  \caption{PF-CBM forward graph. Frozen DINOv3 patch features enter a trainable head. A soft foreground gate produces per-patch probabilities; $P$ part queries cross-attend with the gate injected in log space and the Gaussian spatial prior added to the logits; a separate GlobalTokenAttention module ($G{=}1$ global query, no spatial prior) covers whole-object global concepts; part tokens and the global token feed part-routed and globally-routed attribute heads whose logits pass through per-concept temperatures and a sigmoid; the linear classifier reads the resulting concept probability vector.}
  \label{fig:arch}
\end{figure}

The input is a tensor of cached DINOv3 patch features $\mathbf{X}\in\R^{B\times N\times D}$, where $B$ is the batch size, $N=1024$ is the number of patch tokens on a $32\times 32$ grid, and $D=768$ is the feature dimension of a ViT-B backbone. Let $P=12$ be the number of anatomical parts, $C=312$ the number of attributes, $K=200$ the number of classes, and $d=384$ the attention inner dimension. A precomputed grid $\bm{\Phi}\in\R^{N\times 2}$ holds the $(\mathrm{row},\mathrm{col})$ patch coordinate of every token, so the model operates in patch units rather than pixels. A forward pass runs four stages, summarized in Figure~\ref{fig:arch}: a foreground gate, part cross-attention, part-routed attribute heads, and a linear classifier on concept probabilities.

\subsection{Foreground gating with DINOv3 features}
\label{sec:gate}
The foreground gate is a small MLP over patch features. Given $\mathbf{X}$ it produces per-patch logits $\mathbf{z}^{\mathrm{fg}}\in\R^{B\times N}$ and the gate $\mathbf{g}=\sigmoid(\mathbf{z}^{\mathrm{fg}})\in[0,1]^{B\times N}$, where $\sigmoid(\cdot)$ is the elementwise logistic function:
\begin{equation}
  \mathbf{H}=\mathrm{GELU}\!\big(\mathrm{LayerNorm}(\mathbf{X})\,\mathbf{W}^{(1)}+\mathbf{b}^{(1)}\big),\qquad
  \mathbf{z}^{\mathrm{fg}}=\mathbf{H}\,\mathbf{w}^{(2)}+b^{(2)},
\end{equation}
with weights $\mathbf{W}^{(1)}\in\R^{D\times 256}$, $\mathbf{w}^{(2)}\in\R^{256}$ and hidden width $256$. We deliberately keep the gate small, because the bulk of the foreground signal is already close to linearly separable in DINOv3 patch space; the module fits a one-dimensional classifier on top of frozen features rather than training a segmenter. The gate answers a single whole-object question (ie, "is this patch on the bird"), and is supervised by a whole-object target (Section~\ref{sec:loss}). It does not use per-part boxes. Separating this question from the question of where each part sits is what later allows box supervision and keypoint supervision to be removed independently.

\subsection{Part cross-attention and the Gaussian spatial prior}
\label{sec:attn}
A bank of $P$ learnable part queries $\mathbf{Q}_0\in\R^{P\times D}$ is projected to queries $\mathbf{Q}=\mathbf{Q}_0\mathbf{W}_Q\in\R^{P\times d}$, with keys $\mathbf{K}=\mathbf{X}\mathbf{W}_K$ and values $\mathbf{V}=\mathbf{X}\mathbf{W}_V$ in $\R^{B\times N\times d}$, using projection matrices $\mathbf{W}_Q,\mathbf{W}_K,\mathbf{W}_V\in\R^{D\times d}$. The attention logits combine the scaled dot product, the gate injected additively in log space, and a per-part log-Gaussian spatial term:
\begin{equation}
  \mathbf{S}_{b,p,n}=\frac{\mathbf{Q}_p^\top\mathbf{K}_{b,n}}{\sqrt{d}}
  \;+\;\log\!\big(\mathbf{g}_{b,n}+\varepsilon\big)
  \;-\;\underbrace{\frac{\|\bm{\Phi}_n-\bm{\mu}_p\|^2}{2\sigma_p^2}}_{\text{spatial prior}},
  \qquad \varepsilon=10^{-6}.
  \label{eq:scores}
\end{equation}
Here $\mathbf{Q}_p\in\R^d$ is the projected query of part $p$, $\bm{\Phi}_n\in\R^2$ is the grid coordinate of patch $n$, and $\bm{\mu}_p\in\R^2$ and $\sigma_p>0$ are the learnable mean and standard deviation of part $p$'s isotropic 2D Gaussian prior (the log-normalizer $-\log(2\pi\sigma_p^2)$ is constant over patches $n$ and cancels under the softmax, so it is dropped). After the softmax over patches, the log-space gate term is equivalent to multiplying the unnormalized attention weight of patch $n$ by $\mathbf{g}_{b,n}$, so a near-zero gate smoothly suppresses a patch while a near-one gate leaves its score unchanged. This avoids both the indeterminate 0/0 form that arises in the softmax normalizer when all gate values vanish, and the zero-subgradient problem of a binary hard mask, which would otherwise block gradient flow to the gate parameters. The part token and its attention centroid are
\begin{equation}
  \mathbf{A}_{b,p,\cdot}=\softmax_n \mathbf{S}_{b,p,n},\qquad
  \mathbf{T}_{b,p,\cdot}=\mathrm{LayerNorm}\!\Big(\sum_{n}\mathbf{A}_{b,p,n}\mathbf{V}_{b,n,\cdot}\Big),\qquad
  \mathbf{m}_{b,p,\cdot}=\sum_{n}\mathbf{A}_{b,p,n}\,\bm{\Phi}_{n,\cdot}.
  \label{eq:tokens}
\end{equation}
The attention is single-head by design, so each part produces exactly one map and the maps stay comparable across images.

\paragraph{Why the prior is needed.}
With random query initialization the concept loss is permutation-invariant with respect to which query covers which part. Any relabeling of the $P$ queries yields the same loss, so gradient descent has no signal to assign one query to the beak and another to the tail. Queries then collapse toward a common location, usually the object centroid. The spatial prior in Eq.~\ref{eq:scores} breaks this symmetry. Both $\bm{\mu}_p\in\R^2$ and $\log\sigma_p\in\R$ are trainable parameters refined by gradient descent; their gradients are nonzero because the quadratic term connects them to the attention distribution and thus to the concept and classification objectives. Breaking the symmetry is necessary but not sufficient for grounding. The prior spreads the queries to distinct regions, yet with the alignment loss removed the centroids do not settle on the named anatomy (Section~\ref{sec:results}). Localization comes from the alignment term, which regresses each centroid toward its keypoint on the images that carry annotation.

\paragraph{Sparse keypoint initialization.}
The prior is a population prior, not a per-image one. Each mean is initialized offline from the average visible keypoint location of that part over a set of annotated training images,
\begin{equation}
  \bm{\mu}_p=\frac{\sum_{i:\,v_{i,p}>0}\bm{\Phi}^{\mathrm{kp}}_{i,p}}{\sum_{i:\,v_{i,p}>0}1},
\end{equation}
where $\bm{\Phi}^{\mathrm{kp}}_{i,p}\in\R^2$ is the patch-space keypoint of part $p$ in image $i$ and $v_{i,p}\in\{0,1\}$ its visibility flag. Each $\log\sigma_p$ is initialized to $\log 5$, a broad prior that keeps queries flexible while preventing early collapse. The initialization only has to answer a coarse ordinal question, which query starts near the beak versus the tail, so it tolerates extreme sparsity. The prior means are never used as per-image targets, at training or at test time. Grounding itself is supplied by the centroid alignment loss (Section~\ref{sec:loss}), and Section~\ref{sec:results} measures how few annotated images that loss actually needs.

\subsection{Slot-to-part routing and concept scoring}
\label{sec:routing}
Routing is enforced structurally by two fixed buffers loaded from a committed concept-to-part mapping: an index map $\mathtt{c2p}\in\{-1,0,\dots,P-1\}^C$ giving the part that owns each concept ($-1$ marks a global concept), and a global mask. Table~\ref{tab:routing} lists the per-part concept counts and example assignments. For a part-routed concept $c$, the routed feature is a direct copy of the assigned part token,
\begin{equation}
  \mathbf{R}_{b,c,\cdot}=\mathbf{T}_{b,\,\mathtt{c2p}[c],\,\cdot},\qquad c\notin\mathrm{Global},
  \label{eq:route}
\end{equation}
and for a global concept (the $19$ \emph{has\_size} and \emph{has\_shape} attributes) it reads from a dedicated global token produced by a separate \textsc{GlobalTokenAttention} module. This module holds $G{=}1$ learnable global query that cross-attends over all foreground patches with the gate injected in log space but no Gaussian spatial prior, because size and shape attributes describe whole-object properties that need not localize to a specific anatomical region. Each global concept is hard-routed to its assigned global token through a fixed index buffer $\mathtt{g}\in\{0,\dots,G{-}1\}^C$, so the routing is structural for every concept in the model. Because Eq.~\ref{eq:route} is a copy and not a learned projection, perturbing any part token other than $\mathtt{c2p}[c]$ leaves concept $c$ unchanged. A wing-color head cannot read the beak token regardless of training, and a size concept cannot access any part token regardless of training. Concept logits, probabilities, and class logits follow as
\begin{equation}
  \bm{\ell}_{b,c}=\mathbf{R}_{b,c,\cdot}^\top\mathbf{W}^{\mathrm{head}}_{c,\cdot}+b^{\mathrm{head}}_c,\qquad
  \mathbf{p}_{b,c}=\sigmoid\!\big(\bm{\ell}_{b,c}/\tau_c\big),\qquad
  \mathbf{y}_{b,k}=\mathbf{W}^{\mathrm{cls}}_{k,\cdot}\,\mathbf{p}_{b,\cdot}+b^{\mathrm{cls}}_k,
  \label{eq:scoring}
\end{equation}
with per-concept head weights $\mathbf{W}^{\mathrm{head}}_{c,\cdot}\in\R^d$ and bias $b^{\mathrm{head}}_c$, classifier weights $\mathbf{W}^{\mathrm{cls}}\in\R^{K\times C}$ and bias $\mathbf{b}^{\mathrm{cls}}\in\R^K$, and a learned per-concept temperature $\tau_c=\mathrm{clamp}(\exp\lambda_c,0.5,5.0)$ with log-parameter $\lambda_c\in\R$. The class head reads only the concept probabilities $\mathbf{p}$, so the concept vector is a genuine bottleneck and a test-time edit to $\mathbf{p}$ propagates to $\mathbf{y}$ through a fixed linear map.

\begin{table}[t]
  \centering
  \small
  \caption{Slot-to-part routing on CUB. Left: number of attributes routed to each of the 12 anatomical part slots (19 global \emph{has\_size}/\emph{has\_shape} attributes use a dedicated global attention token). Right: example concept-to-part assignments. Each attribute head reads only from the listed part token.}
  \label{tab:routing}
  \begin{minipage}{0.46\linewidth}\centering
  \begin{tabular}{lr lr}
    \toprule
    Part & \# & Part & \# \\
    \midrule
    tail  & 40 & breast   & 19 \\
    wing  & 39 & forehead & 15 \\
    back  & 34 & leg      & 15 \\
    belly & 34 & nape     & 15 \\
    beak  & 27 & throat   & 15 \\
    crown & 26 & eye      & 14 \\
    global & 19 & \textbf{total} & \textbf{312} \\
    \bottomrule
  \end{tabular}
  \end{minipage}\hfill
  \begin{minipage}{0.5\linewidth}\centering
  \begin{tabular}{ll}
    \toprule
    Attribute & Part slot \\
    \midrule
    \texttt{has\_bill\_shape::all-purpose} & beak \\
    \texttt{has\_wing\_color::black}       & wing \\
    \texttt{has\_throat\_color::white}     & throat \\
    \texttt{has\_belly\_pattern::striped}  & belly \\
    \texttt{has\_eye\_color::black}        & eye \\
    \texttt{has\_size::small}              & global \\
    \bottomrule
  \end{tabular}
  \end{minipage}
\end{table}

\subsection{Training objective}
\label{sec:loss}
Training minimizes a composite loss
\begin{equation}
  \mathcal{L}=w_{\mathrm{cls}}\mathcal{L}_{\mathrm{cls}}+w_{\mathrm{cpt}}\mathcal{L}_{\mathrm{cpt}}+w_{\mathrm{fg}}\mathcal{L}_{\mathrm{fg}}+w_{\mathrm{al}}\mathcal{L}_{\mathrm{al}}+w_{\mathrm{ent}}\mathcal{L}_{\mathrm{ent}}.
  \label{eq:loss}
\end{equation}
The concept term $\mathcal{L}_{\mathrm{cpt}}$ is a certainty-weighted binary cross-entropy on the temperature-scaled concept logits, where CUB certainty codes (guess, probably, definitely, definitely-with-reason) become per-example weights $\{0,\tfrac13,\tfrac23,1\}$ and missing labels are excluded. The foreground term $\mathcal{L}_{\mathrm{fg}}$ is a binary cross-entropy on the gate logits $\mathbf{z}^{\mathrm{fg}}$ against a per-patch target in $[0,1]^{B\times N}$. That target is either a patch-level bounding-box mask, or, in the box-free setting, a soft foreground map derived from the principal components of frozen DINOv3 patch features. The PCA prior selects, per image, the component whose activation is most concentrated in a central window relative to the border, then normalizes it into a soft foreground map; it is used only to synthesize a training target and never replaces the learned gate at inference. The alignment term $\mathcal{L}_{\mathrm{al}}$ is a visibility-masked smooth-$L_1$ regression of the attention centroids $\mathbf{m}$ onto available keypoints, and is disabled when keypoints are removed. The entropy term $\mathcal{L}_{\mathrm{ent}}$ penalizes the entropy of each attention map to encourage concentration, with a small weight. Setting an auxiliary weight to zero skips that term and turns each supervision source into a clean ablation.

We train in two stages. Stage 1 ($w_{\mathrm{cls}}=0$) pretrains the gate, part attention, and attribute heads on the concept, foreground, alignment, and entropy terms for 10 epochs. Stage 2 ($w_{\mathrm{cls}}=1$) trains jointly for 90 epochs with two parameter groups: the classifier and temperatures at learning rate $3\times10^{-4}$, and the gate, part attention, and heads at $3\times10^{-5}$. The order-of-magnitude smaller rate on the grounding path prevents classification pressure from overwriting the concept-aligned state from Stage 1; without it the gate collapses to a near-constant map as the classifier finds shortcut correlations.

\section{Experiments}
\label{sec:experiments}

\paragraph{Dataset.}
All our experiments use CUB-200-2011 \citep{wah2011cub}, about $11{,}788$ images of $200$ bird species, with $312$ binary attributes per image (each with a $0$--$4$ certainty code), $15$ part keypoints per image with visibility flags, and a whole-object bounding box. We hold out a class-stratified $10\%$ carve of the training split as validation, and we report metrics on the native test split. We merge the $15$ raw keypoints into the $12$-part basis by averaging the three bilateral pairs (ie, eye, leg, wing).

\paragraph{Baselines and conditions.}
We organize the conditions as a ladder that removes one source of supervision at a time:
\begin{itemize}[leftmargin=1.5em,itemsep=2pt,topsep=2pt]
  \item \emph{baseline}: a box-supervised PF-CBM with box-derived foreground supervision and per-image keypoint alignment, which we treat as the reference system;
  \item \emph{spatial prior}: adds the learnable Gaussian prior on top of box-cropped features, keeping the per-image keypoint alignment;
  \item \emph{box-free}: replaces the box foreground target with the PCA prior on full-image features, in a standard and a tightened variant;
  \item \emph{alignment on/off}: at the box-free spatial-prior configuration, trains with the centroid alignment loss enabled or disabled, which isolates the grounding contribution of per-image keypoints and defines the annotation-budget sweep between the two ends;
  \item \emph{part-identity probes}: either remove the spatial prior and keypoint alignment entirely (ie, \emph{no-kp}), or replace ground-truth keypoints with PCA-derived pseudo-keypoints from $k$-means part prototypes, again in a standard and a tight-gate variant.
\end{itemize}
We add two external points of comparison, the original CBM on the $112$-attribute CUB subset \citep{koh2020concept} and DOT-CBM \citep{xie2025dotcbm} at $85.39\%$ CUB top-1.

\paragraph{Metrics.}
We measure recognition with top-1 and top-5 classification accuracy. Mean per-concept ROC AUC measures concept quality and is insensitive to CUB's label sparsity, where a fixed threshold over-triggers on rare positives; we also report per-concept F1 and 15-bin expected calibration error (ECE). For grounding, we use the pointing game (ie, the fraction of image-part pairs whose argmax-attention patch falls within one patch of the keypoint) and the attention-centroid distance in patch units. Concept-class mutual information flags leakage, and concept-intervention curves measure how the prediction responds to corrected concepts. We treat AUC and pointing as the primary measures, because they separate ranking quality and localization from the threshold and calibration effects that F1 conflates.

\paragraph{Implementation.}
The backbone is a frozen \texttt{dinov3-vitb16} at $512\times512$ input, giving a $32\times32$ patch grid and $D=768$. All our runs use $P=12$, $C=312$ unless noted, attention inner dimension $384$, batch size $256$, gradient clipping at $1.0$, the two-stage schedule above with AdamW and cosine annealing in Stage 2, and seed $0$. We never run the backbone inside the training or evaluation loop; instead, we precompute patch features and all spatial side information into an on-disk cache, so a run iterates only the trainable head.

\section{Results and Discussion}
\label{sec:results}

\subsection{Main results}
Table~\ref{tab:main} reports test-set metrics across all conditions. The spatial-prior rows train with per-image keypoint alignment, and the row labeled \emph{no keypoint alignment} is the true no-per-image-supervision condition, the comparison that actually shows where grounding comes from. On the box-free model (PCA foreground target plus Gaussian prior), turning the centroid alignment loss on lifts pointing accuracy sharply while barely moving top-1 accuracy, and the same pattern holds for the box-supervised prior. The prior alone does not localize the queries; on its own it reaches only a small improvement over a model with no spatial structure at all. What the Gaussian prior does is separate the queries into distinct regions. What moves each one onto its correct anatomical part is the alignment loss. Removing bounding-box supervision entirely and replacing it with the PCA foreground target costs almost nothing on classification. We read the box-free model's strong grounding under alignment as a benefit of the PCA-derived foreground target, which is a cleaner signal for the gate than a box-derived mask and lets the aligned part queries localize more precisely.

\begin{table}[t]
  \centering
  \small
  \caption{Test-set metrics on CUB-200-2011. Point is pointing-game argmax accuracy; Dist.\ is mean attention-centroid distance in patch units (lower is better). Bold marks the best value per column. The spatial-prior conditions come in matched pairs: \emph{kp align} trains the centroid alignment loss on per-image keypoints, while \emph{no kp align} disables it and is the true no-per-image-supervision condition.}
  \label{tab:main}
  \begin{tabular}{llccccc}
    \toprule
    Role & Condition & Top-1 & Top-5 & AUC & Point & Dist.\ $\downarrow$ \\
    \midrule
    Baseline (box + kp) & \texttt{pfcbm} & 88.95 & 98.19 & \textbf{76.33} & 36.38 & 1.074 \\
    Spatial prior, box, kp align & \texttt{spatial\_prior} & \textbf{89.01} & 98.17 & 75.91 & 52.18 & 1.036 \\
    Spatial prior, box, no kp align & \texttt{spatial\_prior\_noalign} & 86.0 & 97.8 & 73.59 & 6.30 & 6.462 \\
    Box-free + PCA & \texttt{no\_bbox\_pca} & 88.26 & 98.17 & 76.00 & 59.16 & 0.797 \\
    Box-free + tight PCA & \texttt{no\_bbox\_pca\_tight} & 88.85 & 98.15 & 76.06 & 59.74 & 0.778 \\
    Box-free + prior, kp align (ours) & \texttt{no\_bbox\_spatial\_prior} & 88.9 & \textbf{98.20} & 75.66 & \textbf{70.40} & \textbf{0.687} \\
    Box-free + prior, no kp align & \texttt{no\_bbox\_spatial\_prior\_noalign} & 86.4 & 97.8 & 73.34 & 8.00 & 4.790 \\
    Ablation: no prior, no kp & \texttt{no\_bbox\_no\_kp} & 84.33 & 97.50 & 72.98 & 2.93 & 5.954 \\
    Replace: pseudo-kp & \texttt{pseudo\_kp} & 85.04 & 97.84 & 72.32 & 10.78 & 6.704 \\
    Replace: pseudo-kp tight & \texttt{pseudo\_kp\_tight} & 86.71 & 97.98 & 73.84 & 21.40 & 4.603 \\
    \bottomrule
  \end{tabular}
\end{table}

Part identity turns out to be the harder form of supervision to remove. Dropping both the bounding box and keypoint alignment, with no spatial prior left to anchor the queries, lowers classification accuracy and collapses pointing accuracy almost entirely, which confirms the permutation-symmetry argument that without any anchor the queries never specialize. Pseudo-keypoints derived from k-means part prototypes, relabeled to named parts by a Hungarian assignment against a small ground-truth subset, recover part of this gap, but they stay well short of the keypoint-supervised runs (Table~\ref{tab:main}). The prototypes are repeatable across images but not reliably semantic, since some attach to correlated background structure such as branches or fences rather than to the bird itself. PCA foreground discovery is enough to find the object, but only partially enough to find its parts.

\subsection{Annotation budget for grounding}
\begin{figure}[t]
  \centering
  \begin{subfigure}{0.5\linewidth}\centering
    \includegraphics[width=\linewidth]{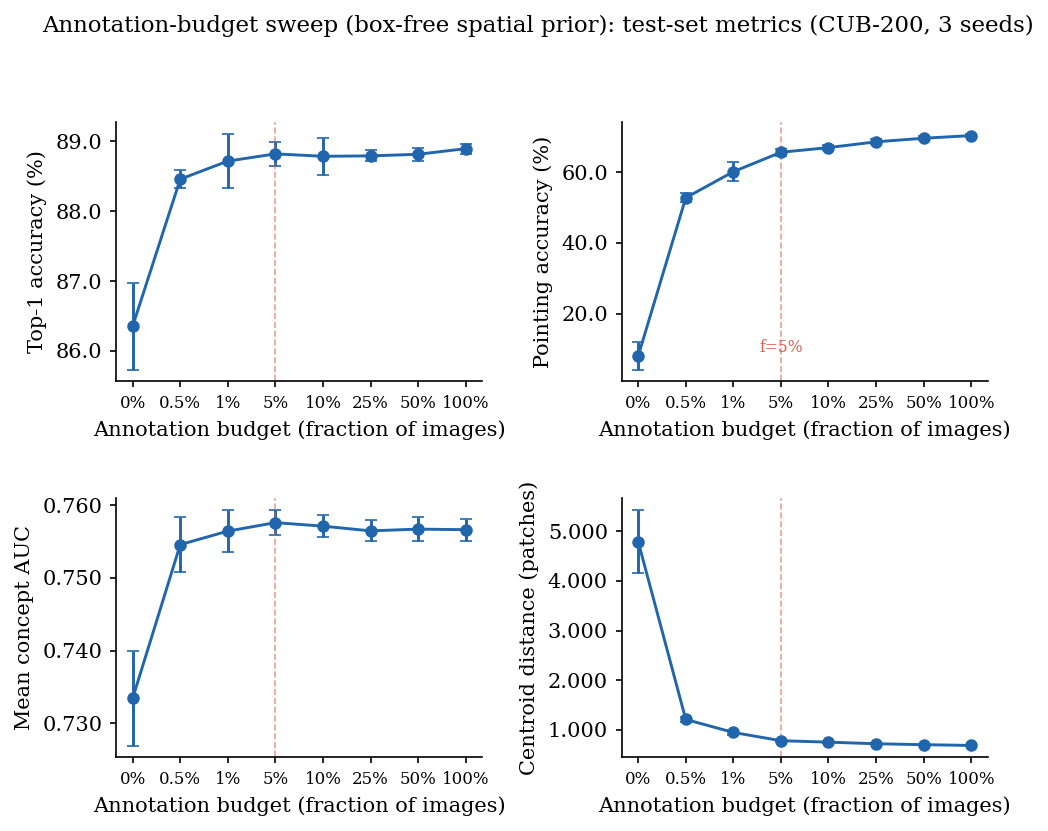}
    \caption{Annotation-budget sweep (box-free).}
    \label{fig:sweep}
  \end{subfigure}\hfill
  \begin{subfigure}{0.46\linewidth}\centering
    \includegraphics[width=\linewidth]{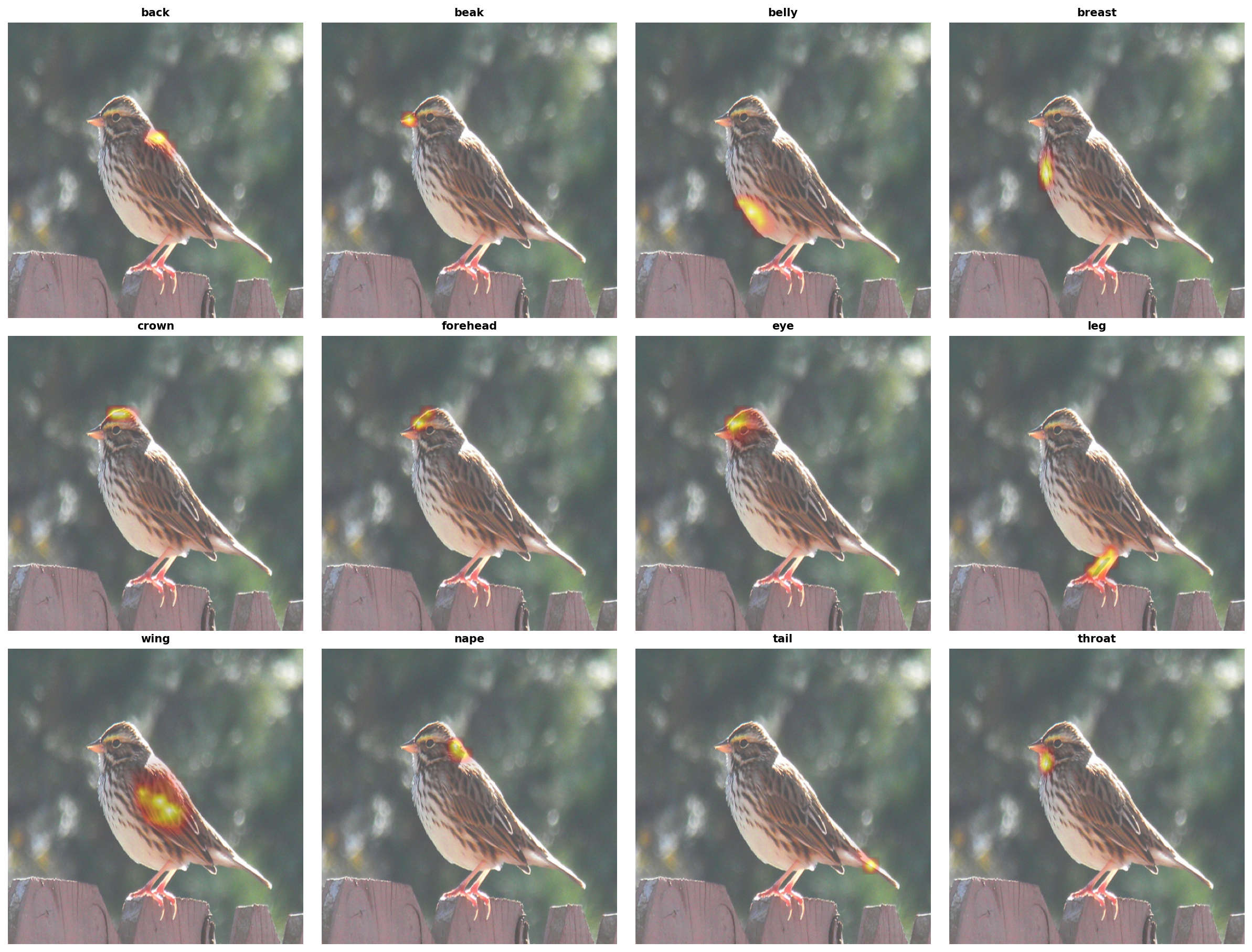}
    \caption{Per-part attention, box-free spatial prior.}
    \label{fig:qual}
  \end{subfigure}
  \caption{(a) Test metrics versus the annotation budget, the fraction of training images whose keypoints supervise the centroid alignment loss (the same subset seeds the prior means), three seeds with $\pm1$ standard-deviation bars. Pointing rises from the $8.0\%$ alignment-off floor to $52.8\%$ at $0.5\%$ of images and saturates near $70.4\%$; top-1 is flat near the baseline throughout. (b) The 12 part attention maps for a held-out image (class 126, correct) from the box-free spatial-prior model, trained with no bounding box and with keypoint alignment: beak, eye, wing, tail, and crown settle on their intended anatomy and the gate suppresses the background post.}
\end{figure}

Grounding comes from the centroid alignment loss, so the natural question is how many annotated images that loss actually needs. We swept an annotation budget from a small fraction of the training set up to full coverage on the box-free spatial-prior model, three seeds per point. At each budget, only the images in that subset contribute their keypoints to the alignment loss, and the same subset seeds the prior means, so a single fraction governs both. Table~\ref{tab:budget} and Figure~\ref{fig:sweep} show the resulting curve. Pointing accuracy rises steeply from the alignment-off floor after just a few dozen annotated images, reaches roughly three-quarters of its full-supervision value with well under one percent of the training set aligned, and then flattens as more images are added. Top-1 accuracy stays essentially flat across the entire sweep and tracks the fully supervised baseline at every budget, so the annotation budget governs grounding quality with no measurable cost to classification. The steep early rise suggests the alignment loss needs only a small, representative set of keypoints to pull each centroid onto its part, and that additional images mainly sharpen the localization rather than change it.

\begin{table}[t]
  \centering
  \small
  \caption{Annotation-budget sweep on the box-free spatial-prior model (PCA foreground plus Gaussian prior). Budget $f$ is the fraction of training images whose keypoints supervise the centroid alignment loss; the same subset seeds the prior means. Three-seed means. The zero-budget row is the alignment-off run.}
  \label{tab:budget}
  \begin{tabular}{lccc}
    \toprule
    Budget $f$ & Approx.\ images & Top-1 & Point \\
    \midrule
    $0\%$ (no alignment) & 0    & 86.4 & 8.0 \\
    $0.5\%$              & 27   & 88.5 & 52.8 \\
    $1\%$                & 54   & 88.7 & 60.1 \\
    $5\%$                & 269  & 88.8 & 65.6 \\
    $10\%$               & 539  & 88.8 & 67.0 \\
    $25\%$               & 1348 & 88.8 & 68.6 \\
    $50\%$               & 2697 & 88.8 & 69.6 \\
    $100\%$              & 5394 & 88.9 & 70.4 \\
    \bottomrule
  \end{tabular}
\end{table}

\subsection{Ablations}
Table~\ref{tab:main} doubles as the ablation over the four mechanisms.

\paragraph{Foreground gating.} Replacing box supervision with the PCA foreground target does not just preserve classification accuracy; it improves pointing accuracy, because the PCA target marks the object more tightly than a rectangular box.

\paragraph{Gaussian prior versus keypoint alignment.} Disabling the centroid alignment loss while keeping the Gaussian prior isolates where grounding actually comes from. Without alignment, the prior alone reaches pointing accuracy only marginally above a model with no spatial structure at all, so the prior breaks the permutation symmetry among the queries but does not localize them. Enabling alignment lifts pointing sharply while leaving top-1 accuracy essentially unchanged. Anatomical localization comes from the alignment loss, not from the prior, which only keeps the queries apart.

\paragraph{Slot-to-part and global routing.} Both routing schemes are always active, since disabling either would change the model class. The structural separation between part tokens, spatially anchored by the Gaussian prior, and the global token, which carries no prior and attends over the full image, is what lets each concept type draw on the evidence appropriate to it by construction. The per-part intervention analysis below shows that correcting the concepts of a single part changes the prediction through that part alone, consistent with this routing guarantee.

\paragraph{Sequential versus joint training.}
Our default trains the concept heads and classifier jointly in Stage 2. We also test a sequential variant that, after the same Stage 1 concept pretraining, replaces joint Stage 2 with a classifier-only stage trained on frozen concept probabilities, so no class gradient reaches the concept encoder. Table~\ref{tab:seq} compares the two. Sequential training gives up a substantial amount of top-1 accuracy at both concept-set sizes we tested, but concept calibration improves sharply and concept-class mutual information drops. The accuracy gap points to how much class-discriminative signal joint training packs into continuous concept magnitudes, information that goes beyond what the binary labels themselves represent. Corrupting half the concepts at random nearly destroys the sequential model's accuracy but barely touches the joint model. The sequential classifier has no path to the class outside the named concepts, so removing half of them removes half its usable information; the joint classifier still has magnitudes to fall back on that binary-level noise does not erase. Grounding follows the same pattern, with joint training also reaching higher pointing accuracy than sequential training. Because sequential training shares only Stage 1 with the joint run, the two variants end up with distinct concept-encoder weights rather than identical ones, so the difference in pointing accuracy reflects a genuine difference in learned attention rather than a change confined to the classifier. We recommend joint training when accuracy or grounding is the priority, and sequential training when the stated concept values must honestly reflect the named attribute, accepting a cost in both accuracy and pointing.

\begin{table}[t]
  \centering
  \small
  \caption{Sequential versus joint training (box-free Gaussian spatial prior, both arms trained with per-image keypoint alignment). C-AUC is mean per-concept AUC, C-ECE mean per-concept calibration error (lower is better), MI mean concept-class mutual information (lower indicates less leakage), Point pointing accuracy. Joint maximizes accuracy and pointing; sequential yields a better-calibrated, genuinely constrained bottleneck.}
  \label{tab:seq}
  \begin{tabular}{lcccccc}
    \toprule
    Model & Top-1 & Top-5 & C-AUC & C-ECE $\downarrow$ & MI $\downarrow$ & Point \\
    \midrule
    Joint-312      & \textbf{89.0} & \textbf{98.2} & 0.757 & 0.221 & 0.660 & \textbf{70.8} \\
    Sequential-312 & 77.1 & 95.2 & \textbf{0.779} & \textbf{0.020} & \textbf{0.321} & 64.2 \\
    \midrule
    Joint-112      & \textbf{82.8} & \textbf{97.0} & 0.797 & 0.213 & 0.716 & \textbf{70.4} \\
    Sequential-112 & 69.7 & 92.2 & \textbf{0.827} & \textbf{0.041} & \textbf{0.639} & 64.1 \\
    \bottomrule
  \end{tabular}
\end{table}

\paragraph{Concept subset and interventions.}
Repeating the box-free spatial-prior run on the $112$-attribute subset of Koh et al.\ \citep{koh2020concept} trades classification accuracy for concept quality. Top-1 accuracy drops because the dropped attributes are disproportionately rare and discriminative, while mean concept AUC and F1 both rise and pointing accuracy holds steady. Routing quality, in other words, does not depend on how many concepts share a part. Because each concept is hard-routed to a single part, correcting only the concepts of one part measures that part's classifier weight directly. Tail, belly, and wing show the deepest gaps under this kind of oracle intervention, tracking their concept counts rather than their grounding quality, and the spatial-prior model matches the supervised baseline closely across all twelve parts. An influence-weighted ordering of concepts causes a faster accuracy decline under intervention than an uncertainty-based ordering, correctly identifying the load-bearing concepts that an expert should inspect first.

\subsection{Qualitative grounding}
Figure~\ref{fig:qual} shows the 12 part attention maps from our box-free spatial-prior model on a held-out image, trained with no bounding box and with per-image keypoint alignment. The maps concentrate on the intended anatomy, the gate suppresses the background post, and the queries do not collapse to a single blob. Because routing is structural, the concept panels surface honest failures: cases where the attention sits on the correct region and the head still calls the wrong color are visibly attributable to the head, not to the model attending elsewhere.

\subsection{Comparison with spatially-grounded CBMs}
\label{sec:comparison_related}

Table~\ref{tab:comparison} situates our PF-CBM among the spatially-grounded CBM methods most directly related to our work. VLG-CBM and DCBM anchor each concept to a detected bounding box or segmentation mask and report concept-quality metrics (ie, ANEC, concept-activation accuracy) as their primary evaluation, so a standard top-1 number is not available from their papers in a directly comparable form. DOT-CBM reports $85.39\%$ top-1 on CUB using a DINOv2 ViT-L/14 encoder; our PF-CBM matches that figure with a smaller ViT-B backbone while also exposing per-part pointing accuracy, a metric that an OT transport plan does not yield as a single interpretable number.

\begin{table}[t]
  \centering
  \small
  \caption{Spatially-grounded CBMs on CUB-200-2011. Supervision lists what per-image annotation is consumed during training. Pointing is the pointing-game accuracy (\%). ``---'' indicates the metric is not reported in a directly comparable form in the source paper. The three PF-CBM rows are the box-free model at different keypoint-alignment budgets (three-seed means): full alignment, alignment on $0.5\%$ of images, and no keypoint alignment.}
  \label{tab:comparison}
  \begin{tabular}{lllcc}
    \toprule
    Method & Grounding & Supervision & Top-1 (\%) & Pointing (\%) \\
    \midrule
    VLG-CBM \citep{vlgcbm2024neurips} & box-anchored   & detector boxes & ---   & --- \\
    DCBM \citep{dcbm2025icml}         & seg-anchored   & seg masks      & ---   & --- \\
    DOT-CBM \citep{xie2025dotcbm}     & soft (OT plan) & saliency prior & 85.39 & --- \\
    \midrule
    PF-CBM, box-free, kp align (ours)     & structural & per-image keypoints  & \textbf{88.9} & \textbf{70.4} \\
    PF-CBM, box-free, $0.5\%$ budget (ours) & structural & ${\sim}27$ kp images & 88.5 & 52.8 \\
    PF-CBM, box-free, no kp align (ours)  & structural & none                 & 86.4 & 8.0 \\
    \bottomrule
  \end{tabular}
\end{table}

DOT-CBM \citep{xie2025dotcbm} and our PF-CBM both ground a CBM in local image evidence using a frozen DINO backbone and a foreground or saliency prior, and they reach comparable CUB accuracy ($85.39\%$ for DOT-CBM with a ViT-L/14 encoder; $88.9\%$ here with a smaller ViT-B). They differ in how grounding is realized. DOT-CBM learns a soft transport plan between patches and concepts, so the patch-to-concept correspondence is an emergent, image-specific optimization that any concept can in principle draw on across patches. Our PF-CBM instead fixes the correspondence at the level of named anatomical parts through a compile-time map, so a concept cannot read outside its assigned part token by construction. Our design gives a guarantee that is trivial to audit and intervention behavior that is predictable from the classifier weights, at the cost of requiring a part vocabulary and a concept-to-part map for the domain. The transport formulation needs neither, but it offers a softer, statistical form of grounding.

\section{Conclusions, Limitations, and Future Work}
\label{sec:conclusion}

We presented a part-factorized concept bottleneck model in which spatial grounding is a structural property of the forward graph. A DINOv3 foreground gate suppresses background, a fixed concept-to-part map routes each attribute to a single anatomical part token, and a learnable Gaussian spatial prior breaks the permutation symmetry among part queries using a dataset-average keypoint initialization. The prior spreads the queries apart but does not localize them; anatomical grounding comes from a centroid alignment loss. On CUB-200-2011, the box-free version of this model reaches classification accuracy close to a fully supervised baseline whether or not it sees per-image keypoints at all. Grounding, however, needs only sparse annotation. Aligning on well under one percent of the training set already recovers roughly three-quarters of the pointing accuracy achieved under full keypoint supervision, and classification accuracy stays nearly flat as the annotation budget grows. Full supervision still gives the best pointing accuracy, so a small budget closes most of the grounding gap but not quite all of it.

The method is best suited to domains where images contain a single object of interest in a reasonably canonical pose, a condition that holds for birds, faces, and cars photographed in standard catalog views, and for similar fine-grained recognition tasks more broadly. Extending it to scenes with arbitrary viewpoints or multiple objects would require object-frame normalization and class-conditional priors to anchor the Gaussian means correctly. A natural next step is an anisotropic or pairwise spatial prior that can represent elongated parts and occlusion, and a stronger naming signal for pseudo-parts, such as text-image matching, would extend the keypoint-free setting to domains that have no keypoint annotation at all.

{\small
\bibliographystyle{plainnat}
\bibliography{references}
}

\clearpage
\appendix

\section{Additional qualitative grounding examples}
\label{app:qual}

This appendix shows five more held-out validation images from the box-free spatial-prior model trained with keypoint alignment, the arm reported at $88.9\%$ top-1 and $70.4\%$ pointing in Table~\ref{tab:main}. For each image we show the 12 part attention maps and the six part-routed concepts predicted with highest probability. Each concept panel gives the concept index, predicted probability, predicted binary label, and CUB ground-truth label. A lime cross marks the ground-truth keypoint for that part when it is visible; a cyan cross marks the model's predicted attention centroid.

\begin{figure}[t]
  \centering
  \includegraphics[width=0.8\linewidth]{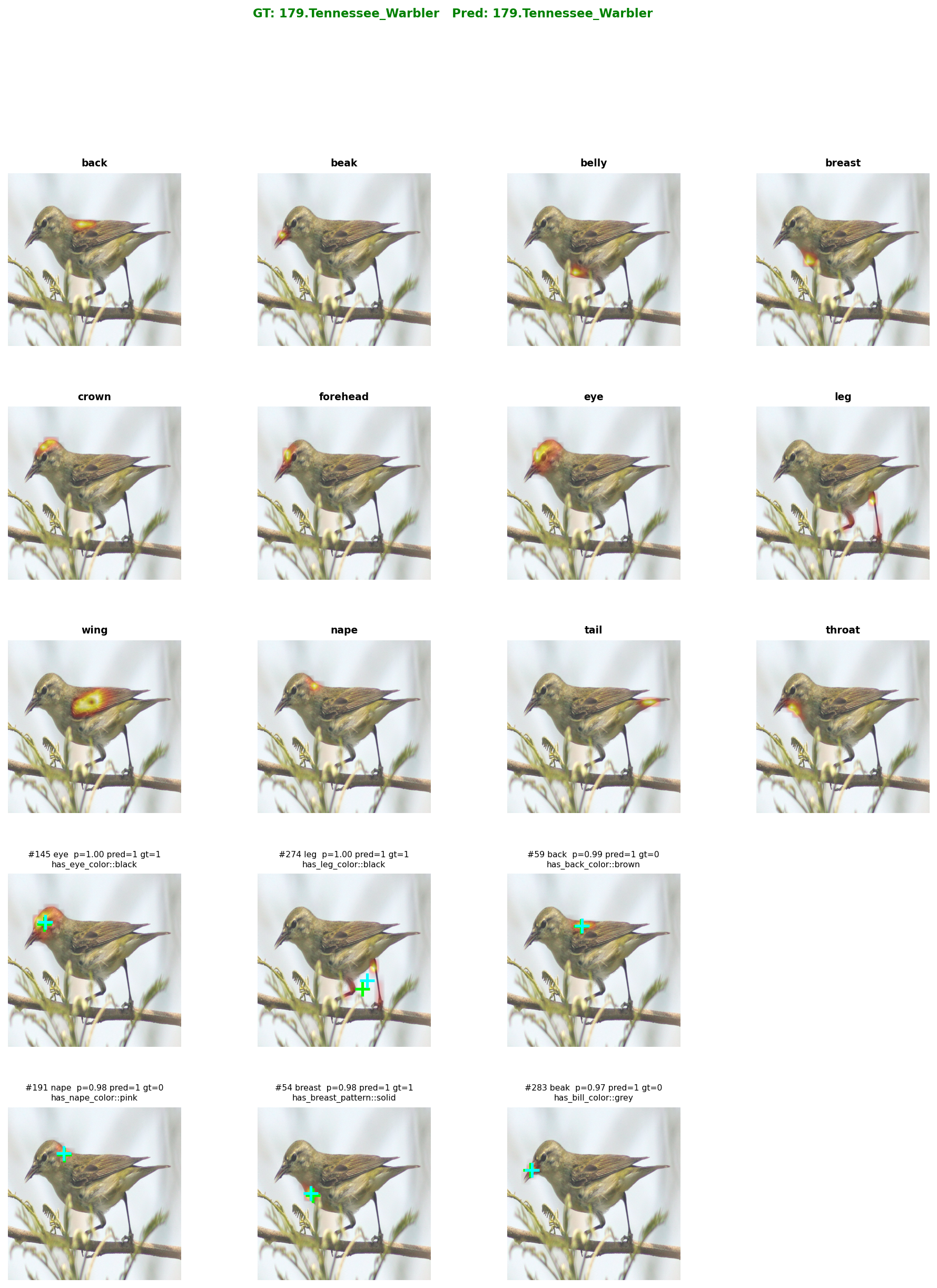}
  \caption{Tennessee Warbler (class 179), correctly classified. Attention separates cleanly across all 12 parts on a small, uniformly olive bird against foliage: eye and forehead land on the head, wing on the folded wing, tail on the tail tip. The top concept panels are mostly correct (\texttt{has\_eye\_color::black}, \texttt{has\_leg\_color::black}), but three panels show a confident wrong call at $p\geq0.97$ where the attention centroid still sits on the right region (back, nape, beak), an honest failure attributable to the color head rather than to grounding.}
  \label{fig:app_ex1}
\end{figure}

\clearpage

\begin{figure}[t]
  \centering
  \includegraphics[width=0.8\linewidth]{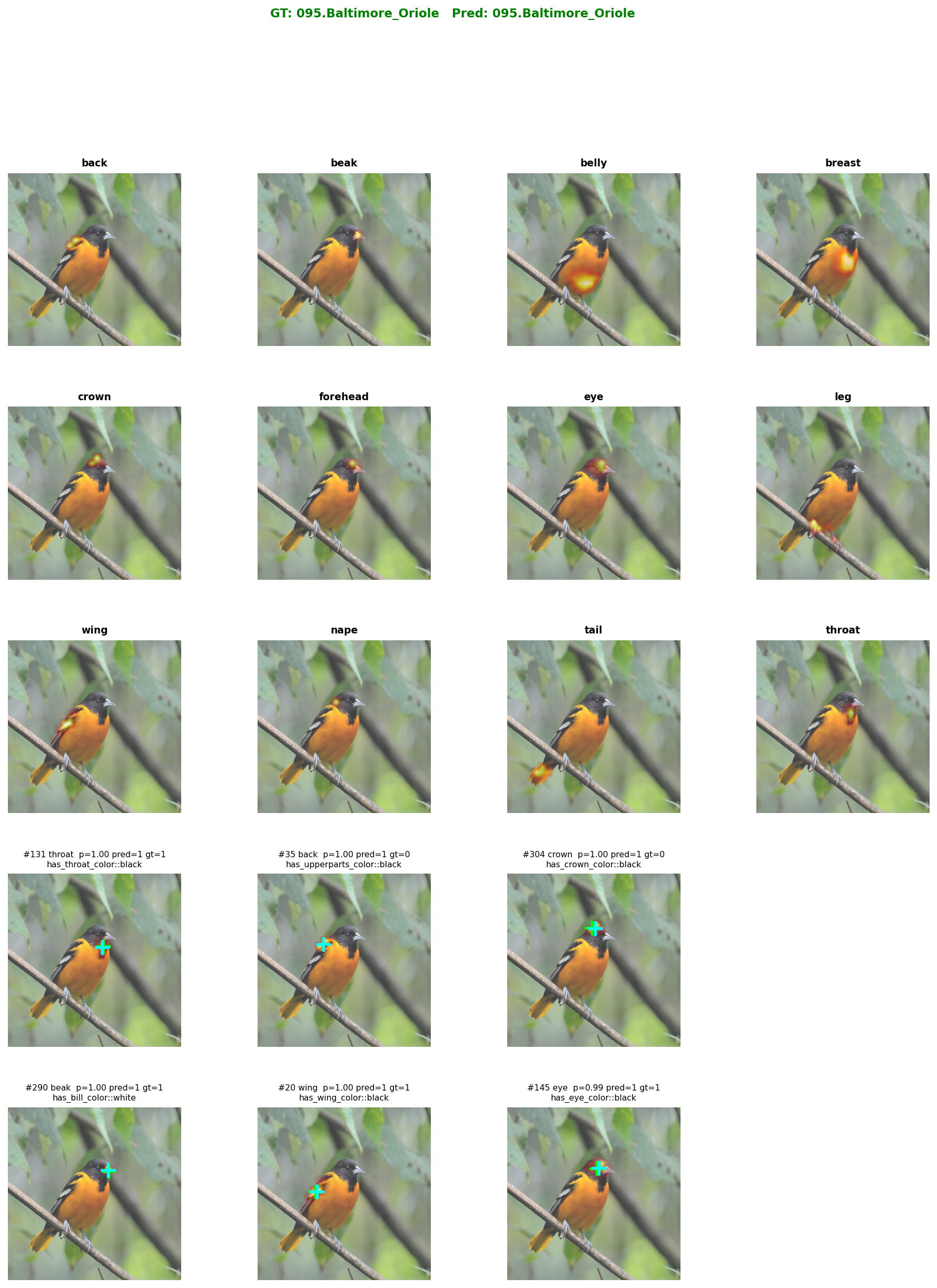}
  \caption{Baltimore Oriole (class 95), correctly classified. The bird's black head and orange body give a strong color contrast, and every part attention lands inside the correct region: throat and crown on the black hood, belly and breast on the orange underparts. \texttt{has\_throat\_color::black}, \texttt{has\_wing\_color::black}, and \texttt{has\_eye\_color::black} all agree with ground truth at $p=1.00$; two panels (\texttt{has\_upperparts\_color::black}, \texttt{has\_crown\_color::black}) are confidently predicted but disagree with the CUB label despite the attention sitting on the correct black plumage, again a color-head error rather than a localization one.}
  \label{fig:app_ex2}
\end{figure}

\clearpage

\begin{figure}[t]
  \centering
  \includegraphics[width=0.8\linewidth]{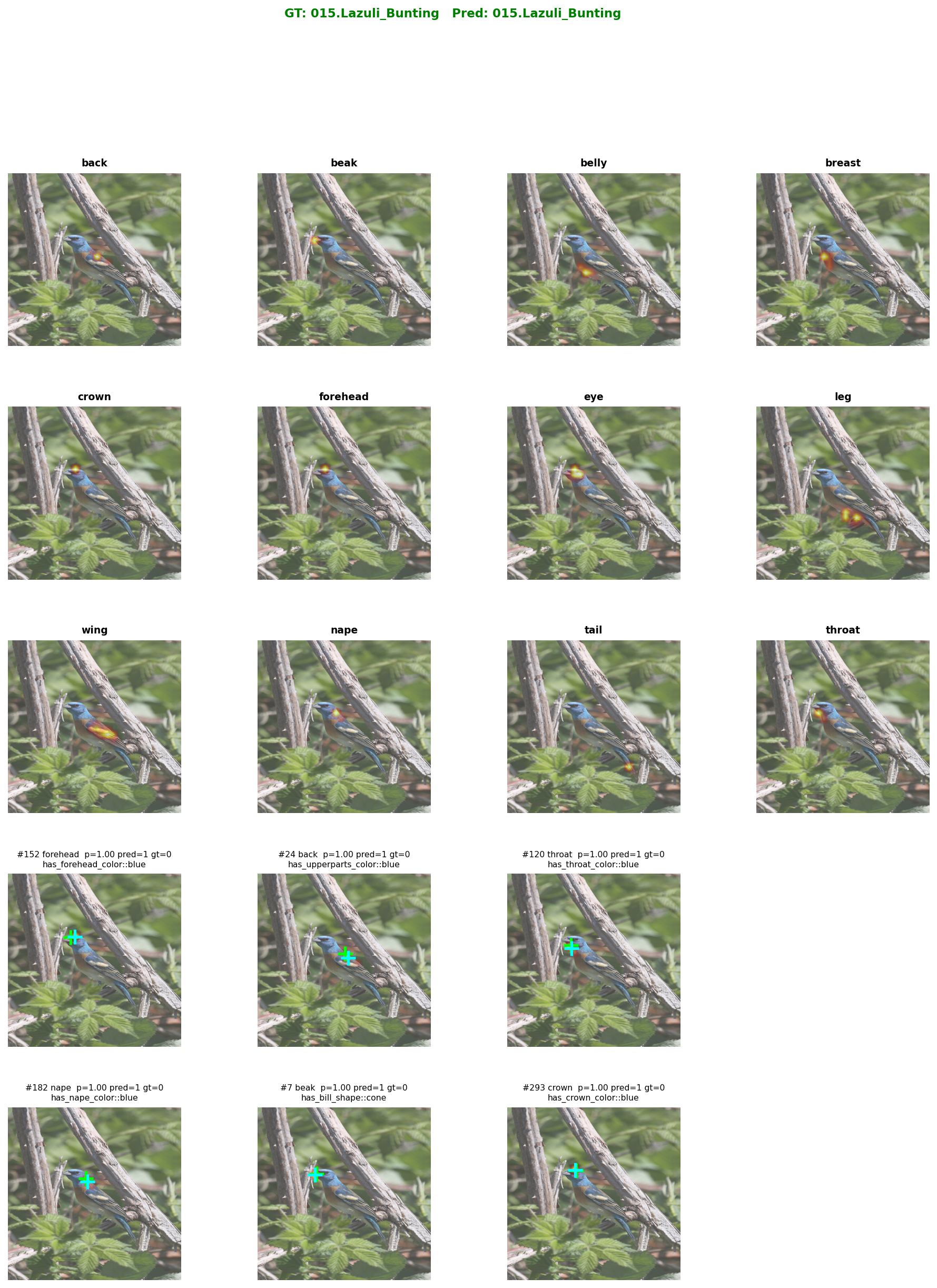}
  \caption{Lazuli Bunting (class 15), correctly classified against a cluttered branch-and-leaf background. All 12 part queries stay tight on the bird's blue body rather than drifting onto the branch. Every one of the top six concept panels predicts a blue attribute at $p=1.00$ (forehead, back, throat, nape, crown) and every one disagrees with the CUB ground-truth label, which records these regions under a different color category for this species. The attention centroids sit correctly on the plumage in each case, so the discrepancy is a labeling or color-vocabulary mismatch rather than a grounding failure.}
  \label{fig:app_ex3}
\end{figure}

\clearpage

\begin{figure}[t]
  \centering
  \includegraphics[width=0.8\linewidth]{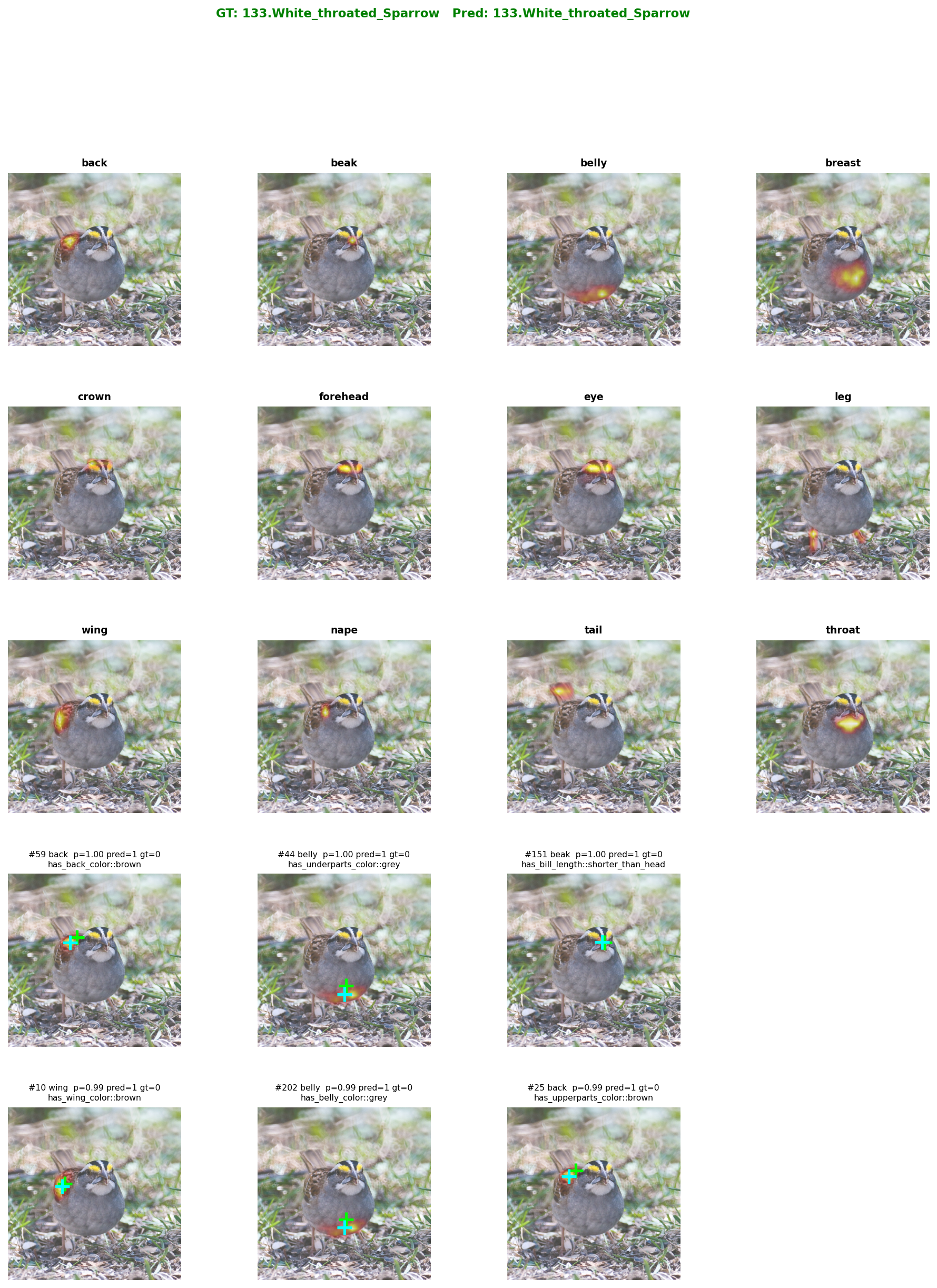}
  \caption{White-throated Sparrow (class 133), correctly classified in a front-facing pose rather than the side profile most training images show. Crown and forehead attention still finds the yellow supraloral patch, throat lands on the white throat patch, and belly and breast separate correctly on the grey underparts. This pose is a harder case for a fixed anatomical query, since several parts (back, tail) are foreshortened or occluded from this angle, yet the queries do not collapse onto a single blob.}
  \label{fig:app_ex4}
\end{figure}

\clearpage

\begin{figure}[t]
  \centering
  \includegraphics[width=0.8\linewidth]{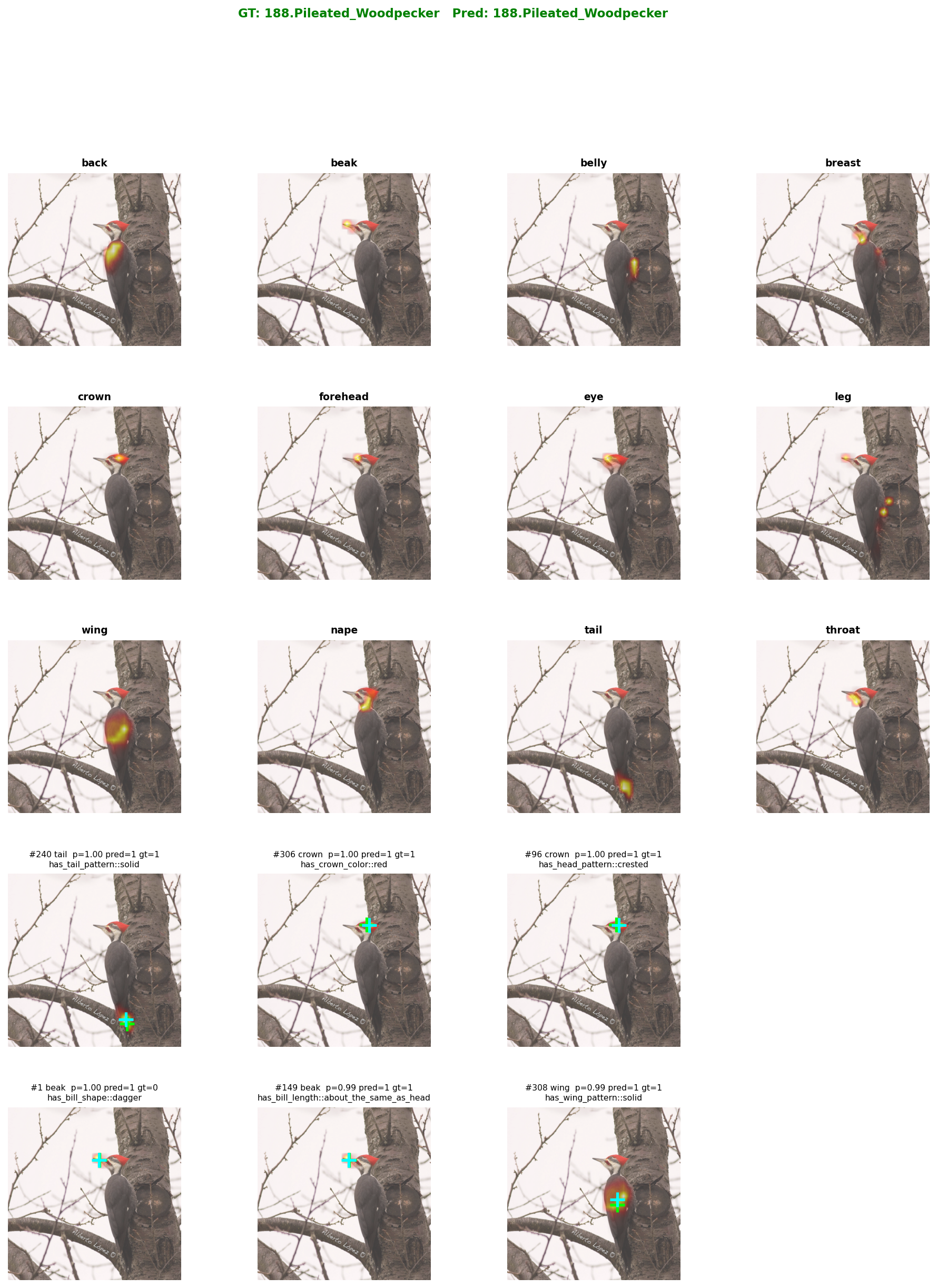}
  \caption{Pileated Woodpecker (class 188), correctly classified on a tree trunk with a heavily textured bark background and a photographer's watermark overlay. Crown attention isolates the red crest precisely, and \texttt{has\_crown\_color::red} and \texttt{has\_head\_pattern::crested} are both correct at $p=1.00$. Beak attention is looser here, spreading partly onto the bark, since the grey bill sits against grey bark of similar texture; this is the one part in the five appendix examples where the background competes visibly with the named region.}
  \label{fig:app_ex5}
\end{figure}

\clearpage

\section{Part attention across the annotation budget}
\label{app:budget_qual}

\begin{figure}[t]
  \centering
  \includegraphics[width=\linewidth]{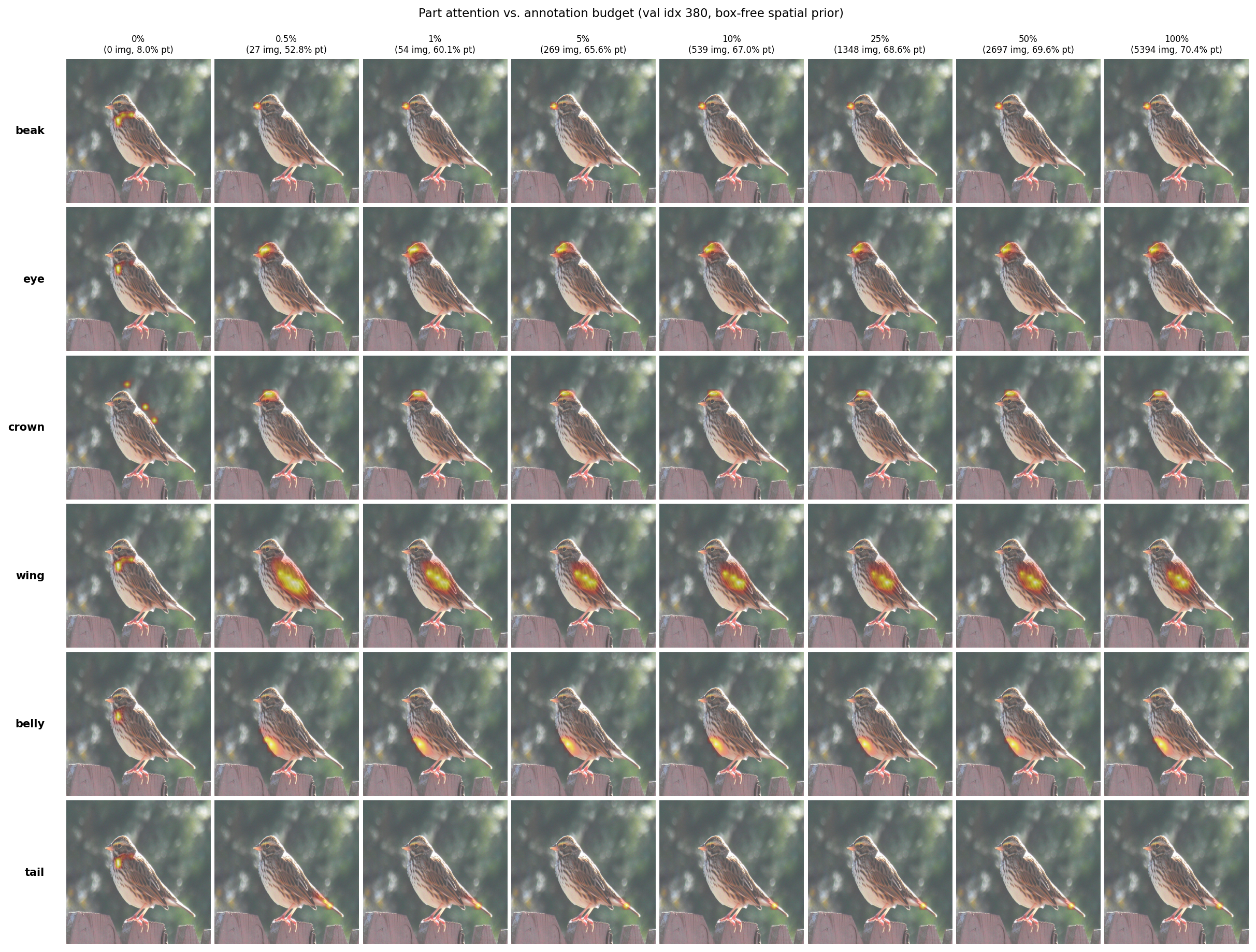}
  \caption{Part attention on one validation image across the annotation budget, from alignment off ($0\%$, $8.0\%$ pointing) to full supervision ($100\%$, $70.4\%$ pointing). Beak and eye are already close to their target region at $0\%$, since the head is salient even without alignment. Wing, belly, and tail are not: at $0\%$, wing attention spreads across most of the body, tail and belly both drift toward the head, and crown splits into two disconnected blobs. From $0.5\%$ (27 images) onward, all six parts snap onto their anatomy and stay stable through $100\%$; the visible change beyond $0.5\%$ is the heatmaps tightening rather than moving, matching the centroid-distance column in Table~\ref{tab:budget}.}
  \label{fig:app_budget_grid}
\end{figure}

Table~\ref{tab:budget} and Figure~\ref{fig:sweep} give the numeric annotation-budget curve. Figure~\ref{fig:app_budget_grid} makes the same curve visible on a single held-out image, the sparrow from Figure~\ref{fig:qual} (validation index 380), by showing six representative part attentions across all eight budgets from the sweep.

\end{document}